\documentclass[runningheads]{llncs}
\usepackage[T1]{fontenc}
\usepackage{graphicx}
\usepackage{hyperref}
\begin{document}
\title{Brain Tumour Removing and Missing Modality Generation using 3D Wavelet Diffusion Model}
\titlerunning{Brain Tumour Removing and Missing Modality Generation using 3D WDM}
\author{André Ferreira \inst{1,2,3,8,9}\orcidID{0000-0002-9332-0091}* \and
Gijs Luijten \inst{3,6} \and
Behrus Puladi \orcidID{0000-0001-5909-6105} \inst{8,9} \and
Jens Kleesiek\inst{3,4,5, 10} \and
Victor Alves\orcidID{0000-0003-1819-7051}\inst{1} \and
Jan Egger\inst{2,3,4,6}
}
\authorrunning{A. Ferreira et al.}
\institute{Center Algoritmi / LASI, University of Minho, Braga, 4710-057,  Portugal \and
Computer Algorithms for Medicine Laboratory, Graz, Austria
\and
Institute for AI in Medicine (IKIM), University Medicine Essen, Girardetstraße 2, Essen, 45131, Germany
\and Cancer Research Center Cologne Essen (CCCE), University Medicine Essen, Hufelandstraße 55, Essen, 45147, Germany
\and German Cancer Consortium (DKTK), Partner Site Essen, Hufelandstraße 55, Essen, 45147, Germany
\and Institute of Computer Graphics and Vision, Graz University of Technology, Inffeldgasse 16, Graz, 8010, Austria
\and Department of Neurosurgery and Spine Surgery, University Hospital Essen, Essen, Germany\\
\and Institute of Medical Informatics, University Hospital RWTH Aachen, Aachen, Germany
\\
\and Department of Oral and Maxillofacial Surgery, University Hospital RWTH Aachen, Aachen,  Germany \\
\and Department of Physics, TU Dortmund University, Dortmund, Germany
\email{*id10656\@alunos.uminho.pt}}
\maketitle              % typeset the header of the contribution
\vspace{-20pt}
\begin{abstract}
%The abstract should briefly summarize the contents of the paper in 150--250 words.
This paper presents the second-placed solution for task 8 and the participation solution for task 7 of BraTS 2024. The adoption of automated brain analysis algorithms to support clinical practice is increasing. However, many of these algorithms struggle with the presence of brain lesions or the absence of certain MRI modalities. The alterations in the brain's morphology leads to high variability and thus poor performance of predictive models that were trained only on healthy brains. The lack of information that is usually provided by some of the missing MRI modalities also reduces the reliability of the prediction models trained with all modalities. In order to improve the performance of these models, we propose the use of conditional 3D wavelet diffusion models. The wavelet transform enabled full-resolution image training and prediction on a GPU with 48 GB VRAM, without patching or downsampling, preserving all information for prediction. The code for these tasks is available at \href{https://github.com/ShadowTwin41/BraTS_2023_2024_solutions}{https://github.com/ShadowTwin41/BraTS\_2023\_2024\_solutions}.

\end{abstract}
\section{Introduction}
Generative models are becoming increasingly explored as these models can significantly optimise productivity and efficiency, shorten development time and improve quality of deep learning solutions. Generative models, more specifically Generative Adversarial Networks (GANs) and Denoising Diffusion Probabilistic Model (DDPM), are the most advanced technologies capable of generating synthetic data indistinguishable from real data. In some cases, DDPMs have outperformed GANs, e.g., generating synthetic natural images \cite{dhariwal2021diffusion}. 

% Talk about inpaint using diffusion models
DDPMs are able to successfully handle tasks such as reconstruction, denoising, image translation, classification among others. They can even be used for tasks in low data regime  \cite{alexis2024diffusion}.  However, these models suffer from some limitations. Due to the use of the Markov chain, they are slow to train and generate new samples. They also require a lot of computational resources, which limits their use. Several efforts have been made to improve the results \cite{nichol2021improved}, reduce the computational effort \cite{friedrich2024wdm}, increase the training speed and reduce the inference time \cite{xiao2021tackling}.

\subsection{Inpainting}
Inpainting is a reconstruction task that has been intensively researched due to the possibility of correcting noisy regions, removing unwanted objects or even replacing a region with an object. Inpainting in natural images has already been extensively researched \cite{lugmayr2022repaint,gsaxner2024deepdr}, but some problems arise when applying it to medical data. The dataset used to train the DDPM may be insufficient to avoid overfitting, leading to poor generations where the generated data may contain multiple artefacts that do not correspond to reality. It could also produce always the same result and therefore not be able to adapt successfully to new cases. In addition, medical data is usually larger than natural images, especially magnetic resonance imaging (MRI) and computed tomography (CT) scans, which have an extra dimensionality, that increases the complexity of training and the amount of resources required.

Since 2023, the Brain Tumour Segmentation (BraTS) Challenge has set a task that challenges the community to present their best algorithm for the inpaiting task. The BraTS datasets contain brain tumours that are properly fully segmented, which is useful for training segmentation models. However, there are a number of algorithms that are not suitable for MRI scans of brains with tumours e.g.  brain parcellation algorithms.
An algorithm capable of removing these lesions would therefore allow these algorithms to be used without modification. As mentioned by the organisers, this would allow a better understanding of the relationship between different brain tumour regions and abnormal brain tissue. Brain modelling would also benefit from these findings \cite{ezhov2021geometry,ezhov2023learn}.

In the 2023 edition, \cite{durrer2024denoising_inpaint} developed a conditional 2D DDPM in which only the slices with the region to be painted over are processed. The model is conditioned by the masked MRI slices and the binary mask in each step of the denoising phase. However, the use of a full 2D DDPM leaded to stripe artifacts.
To solve this problem, \cite{durrer2024denoising} evaluated 6 distinct methods for healthy brain tissue inpainting, namely DDPM 2D slice-wise \cite{durrer2024denoising}, DDPM 2D seq-pos \cite{durrer2024denoising}, DDPM Pseudo3D \cite{zhu2023make}, DDPM 3D mem-eff \cite{bieder2023memory}, LDM 3D \cite{khader2022medical} and 3D WDM \cite{friedrich2024wdm}. The best model (DDPM Pseudo3D) achieved the best results in the test set, i.e. 0.0103, 20.9258, 0.8527 for mean absolute error (MSE), peak-signal-noise-ratio (PSNR) and similarity index measure (SSIM) respectively.  DDPM Pseudo3D is characterised by the use of 2D convolutional layers followed by 1D convolutions in the z-axis, which allows for better continuity between layers and results in an efficient pseudo-3D network. All networks can be trained on a GPU with 48 GB VRAM, with the exception of DDPM 3D mem-eff, which requires more than 78 GB and does not achieve the same results as DDPM Pseudo3D. 3D WDM is one of the networks capable of processing the image at full resolution with the lowest sampling time and memory consumption. However, this network produced very poor results (0.1060, 10.5693, 0.6074 for MSE, PSNR and SSIM respectively) probably due to the chosen pipeline for training and inpainting as well as the short training time.

Zhu et al. \cite{zhu2024advancing} suggests the use of three distinct networks, namely pGAN \cite{dar2019image}, ResViT \cite{dalmaz2022resvit},
and Palette \cite{saharia2022palette}. pGAN and ResViT are conditional GANs and Palette is a conditional diffusion model. Although these networks were adapted and were able to achieve good results, they were not able to achieve better results than the baseline (3D Pix2Pix \cite{isola2017image}).

\subsection{Missing MRI}
MRI scans are one of the main imaging technique for detection and segmentation of brain tumours. Generally, four modalities are acquired in order to produce the best clinical decision, namely T1-weighted with and without contrast enhancement, T2-weighted images, and FLAIR. Several automatic tools require the input of all modalities to work properly.

However, obtaining MRI data is very time-consuming and expensive. Patients who suffer from claustrophobia, have a mental disorder, or move too much during the procedure are challenging to scan, often resulting in data with artifacts. For some patients with metal implants, it may even be impossible.

Often, in real clinical practice is difficult to obtain all MRI modalities, which results in missing MRI sequences. Therefore, since 2023 the BraTS challenge also contains a task for Brain MRI Synthesis for Tumour Segmentation (BraSyn). The objective of this task is to generate the missing modality having the other three as reference. For this, researchers are invited to develop automatic image-to-image translation tools. With such a tool, it is possible to reduce acquisition time, or generate the missing modality to use in other algorithms to improve the detection and segmentation.

Pix2Pix \cite{isola2017image} is one of the most frequently used approaches to fulfil this task. \cite{yu2019ea} used an edge-aware GAN (Ea-GANs) which integrated edge information to reduce the slice discontinuity
and increase sharpness, which are problematic in several medical tasks that use conditional GAN models. Ea-GANs, which is based on the Pix2Pix model, uses T1 scans to generate T2 and FLAIR scans.

In 2023 BraTS edition, \cite{baltruschat2024brasyn} explored the use of Pix2Pix optimised with several distinct loss functions. Loss functions based on pixel-to-pixel similarities (e.g., mean absolute error and similar), adversarial loss, structural SSIM, frequency domain consistency, and latent feature (VGG-based perceptual) consistency. The submitted solution was the combination of all losses, which resulted in the best solution among the participating teams.

We propose the use of conditional 3D WDM \cite{friedrich2024wdm} for tasks 7  BraTS Synthesis (Global) - Missing MRI \cite{li2023brain} and 8 Synthesis (Local) - Inpainting \cite{kofler2023brain}. Our experiments have shown that using several different healthy regions for training, and a distinct scheduler and pipeline for sampling resulted in a strong and efficient solution.

\section{Materials and methods}
Two machines were used for the experiments. The first machine is an IKIM cluster node with 6 NVIDIA RTX 6000, 48 GB of VRAM, 1024 GB of RAM, and AMD EPYC 7402 24-Core Processor. The second is a RWTH aachen cluster with NVIDIA H100 GPUs with 96 GB of VRAM, 512GB of RAM and CPUs Intel Xeon 8468 Sapphire.  Only one GPU was use per training.

\subsection{Datasets}

\subsubsection{Synthesis (Local) - Inpainting task 8:}
The dataset uses the same cases as the BraTS 2023 adult glioma segmentation dataset \cite{baid2021rsna}. It consists of 1251 cases acquired by several different institutions under standard clinical conditions \cite{menze2014multimodal,bakas2017advancing,bakas2018identifying}. Each case consists of a T1 MRI scan, a mask segmenting the unhealthy tissue based on the original segmentation dilated to account for the mass effect, a mask of a healthy region, a full mask and the voided T1 scan, i.e. the T1 scan with the masked region removed. 

The validation and test sets contain only the voided T1 scans and the masks of healthy and unhealthy tissue. The healthy masks are created by selecting a random real unhealthy mask, randomly flipping and rotating it, and placing it in a healthy brain region, as explained in \cite{kofler2023brain}.

\subsubsection{Synthesis (Global) - Missing MRI task 7:}
Similarly to the inpainting task, the BraSyn dataset is based on the BraTS 2023 adult glioma segmentation dataset \cite{baid2021rsna}. The training data contains all four modalities and the respective segmentation. In the validation (219) and test sets (570 cases), a random modality is omitted. The objective is to generate the missing modality.

\subsection{3D wavelet diffusion model}

3D WDM \cite{friedrich2024wdm} uses the wavelet transform to reduce the spatial dimensionality by a factor of 8, splitting the image into eight different channels. This enables the use of a single graphics processor with 48 GB VRAM to process the scans in full resolution. 

\subsubsection{Synthesis (Local) - Inpainting task:}

\begin{figure}
\includegraphics[width=\textwidth]{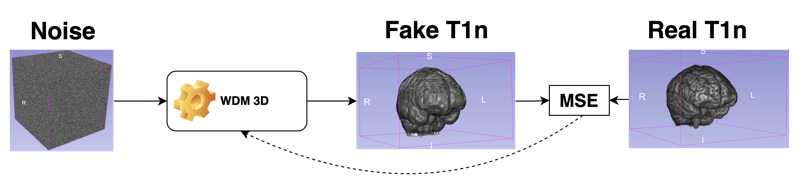}
\caption{Training pipeline of the Default model} \label{fig:Default_model}
\end{figure}

\begin{figure}
\includegraphics[width=\textwidth]{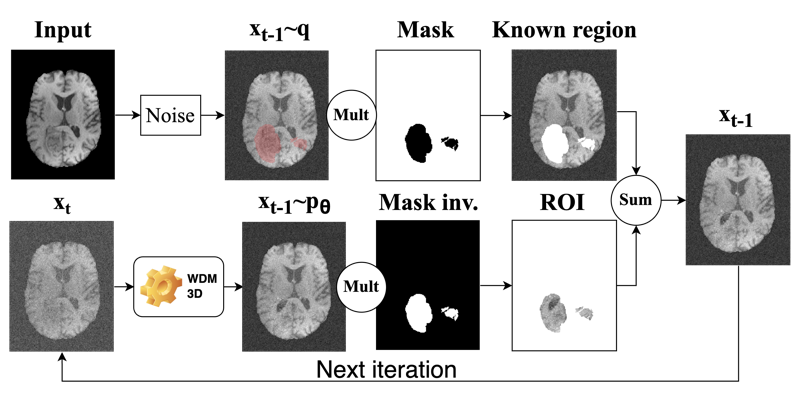}
\caption{Conditional sampling pipeline.} \label{fig:Conditional_sampling}
\end{figure}

As a baseline model, we train a 3D WDM model without any changes, as shown in Figure \ref{fig:Default_model}. The loss function is formally defined by equation \ref{eq:loss_D}. This model is referred to as "\textit{Default ($D_l$)}". For conditional sampling, we use the strategy explored in \cite{lugmayr2022repaint}, which samples the real scan for each time $t-1$ and replaces the region of interest (ROI) with the prediction of the model (as represented in Figure \ref{fig:Conditional_sampling}). 

\begin{equation}
\label{eq:loss_D}
L_{D_l} = L_{MSE} = ||\tilde{x}_0-x_0||^2_2
\end{equation}
where MSE is the mean squared error (MSE), $\tilde{x}_0$ is the predicted scan and $x_0$ the real scan.

\begin{figure}
\includegraphics[width=\textwidth]{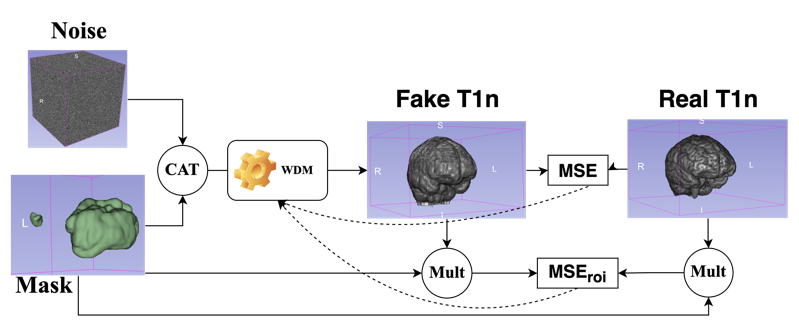}
\caption{Training pipeline of the Conditional Default model} \label{fig:Conditional_Default_model}
\end{figure}

We train a model that takes both the scan and the mask (both unhealthy and healthy masks) as input to give more information to the network and make it conditional as shown in Figure \ref{fig:Conditional_Default_model}. The loss is composed of the MSE of the overall scan and the MSE of the reconstructed region, as formally defined in equation \ref{eq:loss_DC} with $\lambda_1=10$.  This model is referred to as "\textit{Default conditional ($DC_l$)}".

\begin{equation}
\label{eq:loss_DC}
L_{DC_l} = L_{MSE} + L_{MSE_{ROI}} = ||\tilde{x}_0-x_0||^2_2 + \lambda_1 ||\tilde{x}_0\cdot m - x_0 \cdot m||^2_2
\end{equation}

where $\tilde{x}_0\cdot m$ is the predicted ROI, i.e., region to inpaint, and $x_0 \cdot m$ the real ROI.

\begin{figure}
\includegraphics[width=\textwidth]{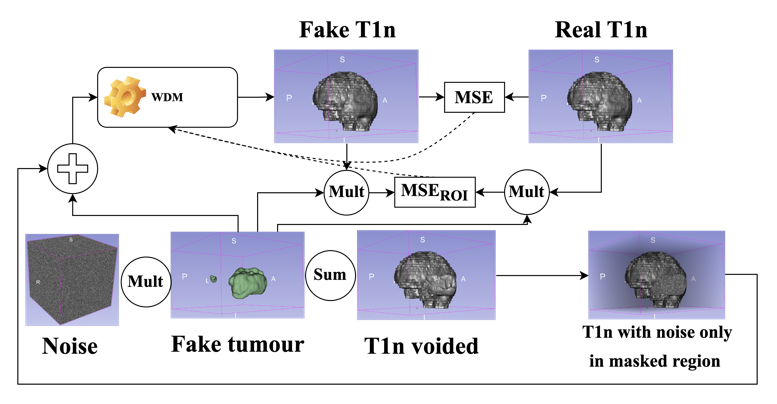}
\caption{Training pipeline of the Conditional Always known model} \label{fig:AK_model}
\end{figure}

We have also trained a model that contains no noise in the "known region", i.e. only the region to be inpainted is noisy. The training pipeline is very similar to the one illustrated in Figure \ref{fig:Conditional_Default_model}, but instead of adding noise to the entire scan, only the region to be inpainted contains noise, as shown in Figure \ref{fig:AK_model}. The sampling pipeline is very similar to the one shown in Figure \ref{fig:Conditional_sampling}, but instead of $x_{t-1}\sim q$ we use voided MRI scans directly. This model also uses both the scan and the corresponding mask, but only the healthy mask. The healthy mask is used to calculate the loss, as the loss is composed of the MSE of the entire scan and the MSE of the reconstructed volume, as formally defined in equation \ref{eq:loss_AK} with $\lambda_1=10$. This model is referred to as "\textit{Always known ($AK_l$)}".

\begin{equation}
\label{eq:loss_AK}
L_{AK_l} = ||\tilde{x}_0-x_0||^2_2 + \lambda_1 ||\tilde{x}_0\cdot m_h - x_0 \cdot m_h||^2_2
\end{equation}
where $m_h$ is the heathy mask.

Lastly, we have trained a model similar to \textit{$AK_l$}, but we remove the unhealthy region of the scan. Therefore, the model only has access to the healthy tissue, computing the loss using only healthy tissue. This model is called "\textit{Always known healthy ($AKH_l$)}". The loss function is formally defined by the equation \ref{eq:loss_AKH} with $\lambda_1=10$ and where $m_{uh}$ is the unhealthy mask. The sampling procedure is the same as for model \textit{$AK_l$}.

\begin{equation}
\label{eq:loss_AKH}
L_{AKH_l} = ||\tilde{x}_0\cdot (1-m_{uh})-x_0\cdot (1-m_{uh})||^2_2 + \lambda_1 ||\tilde{x}_0\cdot m_h - x_0 \cdot m_h||^2_2
\end{equation}

For training, we selected Time=1000 and noise\_scheduler=linear. The number of iterations per model will be described in the results section. For inpainting, we tested the use of several different schedulers: Time=[1000,2000,3000,4000,5000] and noise\_scheduler=[linear,cosine,DPM++ 2M Karras]. The remaining hyperparameters were not changed from the original experiment \cite{friedrich2024wdm}.

\subsubsection{Synthesis (Global) - Missing MRI task:}

As a baseline for this task, we also trained a 3D WDM without any modifications, except in the number of channels of input and output, instead of using one channel, four channels were used, as the objective was to compute the four modalities with the baseline. This model will be called "Default ($D_g$)". The loss function is formally characterised by equation \ref{eq:loss_D_g}. For the prediction of one modality, the other three are sampled at every step $t-1$ from the forward process and used as input of the model jointly with the predicted modality.

\begin{equation}
\label{eq:loss_D_g}
L_{D_g} = L_{MSE} = ||\tilde{a}_0-a_0||^2_2
\end{equation}
where $\tilde{a}_0$ are the four modalities generated and $a_0$ the real four modalities.

To further improve our solution we explored two distinct pipelines for training and sampling:

\textbf{Known all time ($KAT_g$):} instead of having the four modalities starting as noise, only the modality to predict has noise, and the remaining three do not have any noise. The sampling method is similar to $D_g$, but instead of adding noise to the known modalities, these are used without any noise to condition the sampling. 

\textbf{Known 3 to 1 ($K3T1_g$):} The second takes only the three known modalities as input and outputs only the missing modality for the training process. The sampling process is very similar as the three known modalities are fed without any noise, and the unknown modality is progressively generated. 

Both models contain four additional channels concatenated after applying the wavelet transformation, in order to inform the network which modality is missing. The channel with modality missing contains only values 1 and the remaining 0. This is specially important for the $K3T1_g$.

\subsection{Data augmentation}
For the task of missing modality, we did not create new scans for training, nor did we use conventional data augmentation such as rotation, mirroring, etc.

For the inpainting task, we use a similar strategy to \cite{kofler2023brain} to increase the number of different masks used as input. However, we had two sources for masks. The first is the real masks from the dataset. The other is a random mask generator. Following a similar approach to \cite{ferreira2024we}, we trained an alpha-GAN based on \cite{ferreira2022generation} to learn the distribution of healthy masks and generate new ones.  

We created new healthy masks with 1, 2 or 3 regions per case. The number of regions is selected semi-randomly, with a probability of 0.45, 0.45 and 0.1 for each event, respectively. The probability of selecting a real healthy mask from the data set or creating a new mask with the mask generator is equal. The geometric centre of the new masks was chosen randomly, making sure that the centre was outside the unhealthy mask and inside the brain. The new masks were not allowed to overlap with unhealthy tissue and had to be within the MRI scan. A total of 38709 masks were created. These new masks were used to train the $DC_l$ and $AK_l$ models.

For our last model ($AKH_l$), we decided to also use the unhealthy masks to add more variability to the dataset. Thus, we move the unhealthy mask to a region where the tissue is healthy and save both masks as healthy, removing the unhealthy tissue, i.e. the unhealthy mask shifted and the healthy mask created a new healthy mask. This way, 5146 new masks were created. A total of 43855 masks were used for the training of $AKH_l$.

\subsection{Evaluation}
\subsubsection{Synthesis (Local) - Inpainting task:}
The performance of the solutions is measured by the structural similarity index (SSIM), peak signal-to-noise ratio (PSNR) and mean square error (MSE), only in the healthy region as no baseline data is available in the unhealthy region.

\subsubsection{Synthesis (Global) - Missing MRI task:}
Two distinct aspects are evaluated. The first image quality where the SSIM will be used. The SSIM will be computed for the healthy part of the brain and for the tumour. Second, it will be evaluated the contribution of the generated modality in a segmentation algorithm, using the Dice score.

\section{Results}

\begin{figure}
\includegraphics[width=\textwidth]{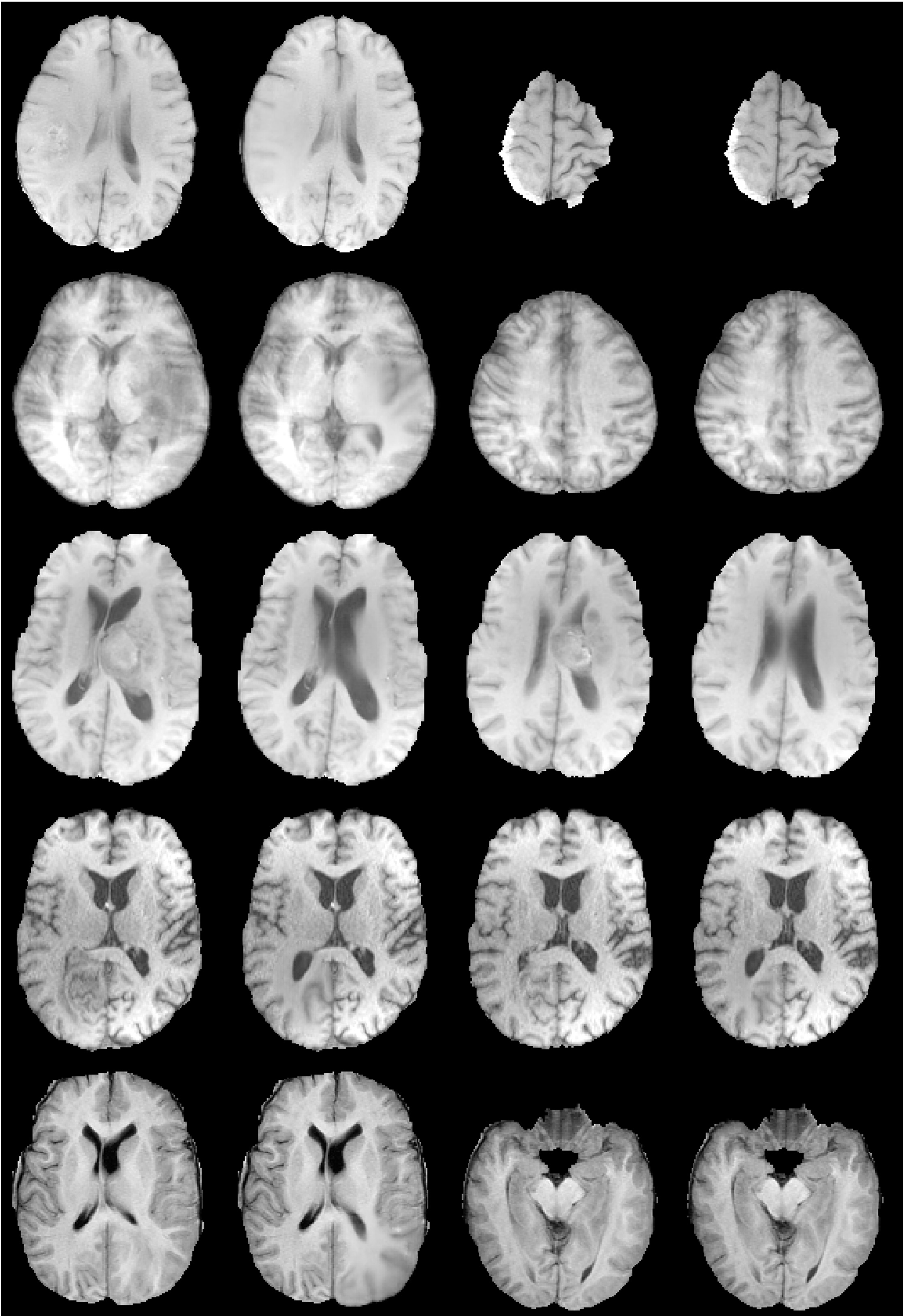}
\caption{Task 8 results. Results of the inpainting model for the training cases 00084-000, 00723-000, 01163-000, 01274-000, 01502-000 in each row. The first column shows a slice of the real case with a tumour, the second the replacement of the tumour with healthy tissue, the third a random slice where the healthy tissue has been removed and the last the respective inpainted slice.} \label{fig:task_8_results}
\end{figure}

Table \ref{tab:results_task8} presents the results for the validation set of the inpainting task conducted on the online platform. Tables \ref{tab:results_task7_test} and \ref{tab:results_task8_test} display the results for the test set as evaluated by the organizers, for tasks 7 and 8, respectively. Since all data was used to train the models at once, the results for the training set are meaningless. The methods are described by their initials and the Time used for scheduling, e.g. $AK_l1000$ means "Always known local task with Time=1000". We do not show the results for cosine scheduler as the visual results were too blurred and very unrealistic compared to linear scheduler. In the testing set, our final solution, i.e., $AKH_l5000$ trained for 3500000 iterations, achieved 0.07±0.04, 22.8±4.41 and 0.91±0.15 for MSE, PSNR and SSIM respectively. Figure \ref{fig:task_8_results} show the results of this solution applied to several training cases.

\begin{table}
\centering
\caption{Results in the validation set using the challenge's online platform. The best results are in bold and the second best underlined. * means that the model was trained for longer, i.e., $AKH_l3000$ and the first $AKH_l5000$ for 3125000 iterations and the second $AKH_l5000$ for 3500000. The remaining were trained for 3000000 iterations.}\label{tab:results_task8}
\begin{tabular}{|c|c|c|c|}
\hline
\textbf{Method} &  \textbf{MSE} & \textbf{PSNR} & \textbf{SSIM}\\
\hline
% Default
$D_l1000$ & 0.033976 ± 0.025657 & 15.83677 ± 3.34655 &  0.689602 ± 0.157043 \\ 
$D_l2000$ & 0.029174 ± 0.025586 & 16.82357 ± 3.72328 & 0.737160 ± 0.143272 \\			
$D_l3000$ & 0.029022 ± 0.026080 & 16.94370 ± 3.86508 & 0.744720 ± 0.142703 \\
% Conditional Default			
$DC_l1000$ & 0.057733 ± 0.036330 & 13.49242 ± 3.60295 & 0.622573 ± 0.178613 \\			
$DC_l2000$ & 0.034858 ± 0.022367 & 15.66606 ± 3.45351 & 0.699688 ± 0.154313 \\					
$DC_l3000$ & 0.031013 ± 0.020530 & 16.28151 ± 3.65336 & 0.720479 ± 0.147351 \\
% Always known
$AK_l1000$ & 0.031369 ± 0.023041 & 16.33596 ± 3.67045 & 0.688697 ± 0.164519 \\
$AK_l2000$ & 0.017569 ± 0.015026 & 18.89508 ± 3.55482 & 0.773430 ± 0.128778\\
$AK_l3000$ &  0.014378 ± 0.010260 & 19.53926 ± 3.34115 & 0.790668 ± 0.119865 \\
$AKH_l3000$ & 0.012153 ± 0.007908	& \underline{20.27791} ± 3.50154 & 0.804382 ± 0.112031 \\
$AKH_l5000$ & 0.010901 ± 0.007649	& \underline{20.75135} ± 3.40085 &	\underline{0.807841} ± 0.114229 \\		
$AKH_l5000$* & \textbf{0.007381} ± 0.005027	& \textbf{22.61126} ± 3.82532 &	\textbf{0.842279} ± 0.098535 \\	

\hline
\end{tabular}
\end{table}

\begin{table}
\centering
\caption{Test set results for task 7 computed by the challenge organisers. The submited solution was $K3T1_g$ with 3000 sampling steps. The DSC are the legacy scores, i.e., does not penalise false negatives and false positives.}\label{tab:results_task7_test}
\begin{tabular}{|c|c|c|c|}
\hline
\textbf{Metric} &  \textbf{ET} & \textbf{TC} & \textbf{WT}\\
\hline
DSC &
0.6808 ± 0.3120 &
0.7301 ± 0.3118 &
0.8345 ± 0.1932 \\
\hline
HD95 &
32.313 ± 93.238 &
30.808 ± 85.888 &
17.746 ± 49.277 \\ 
\hline
SSIM &
\multicolumn{3}{|c|}{ 0.8172 ± 0.0193} \\ 
\hline
\end{tabular}
\end{table}

\begin{table}
\centering
\caption{Test set results for task 8 computed by the challenge organisers.}\label{tab:results_task8_test}
\begin{tabular}{|c|c|c|c|}
\hline
\textbf{Method} &  \textbf{MSE} & \textbf{PSNR} & \textbf{SSIM}\\
\hline
% Default
$AKH_l5000$ & 0.07 ± 0.04 & 22.8 ± 4.41 &  0.91 ± 0.15 \\ 
\hline
\end{tabular}
\end{table}

\begin{figure}
\includegraphics[width=\textwidth]{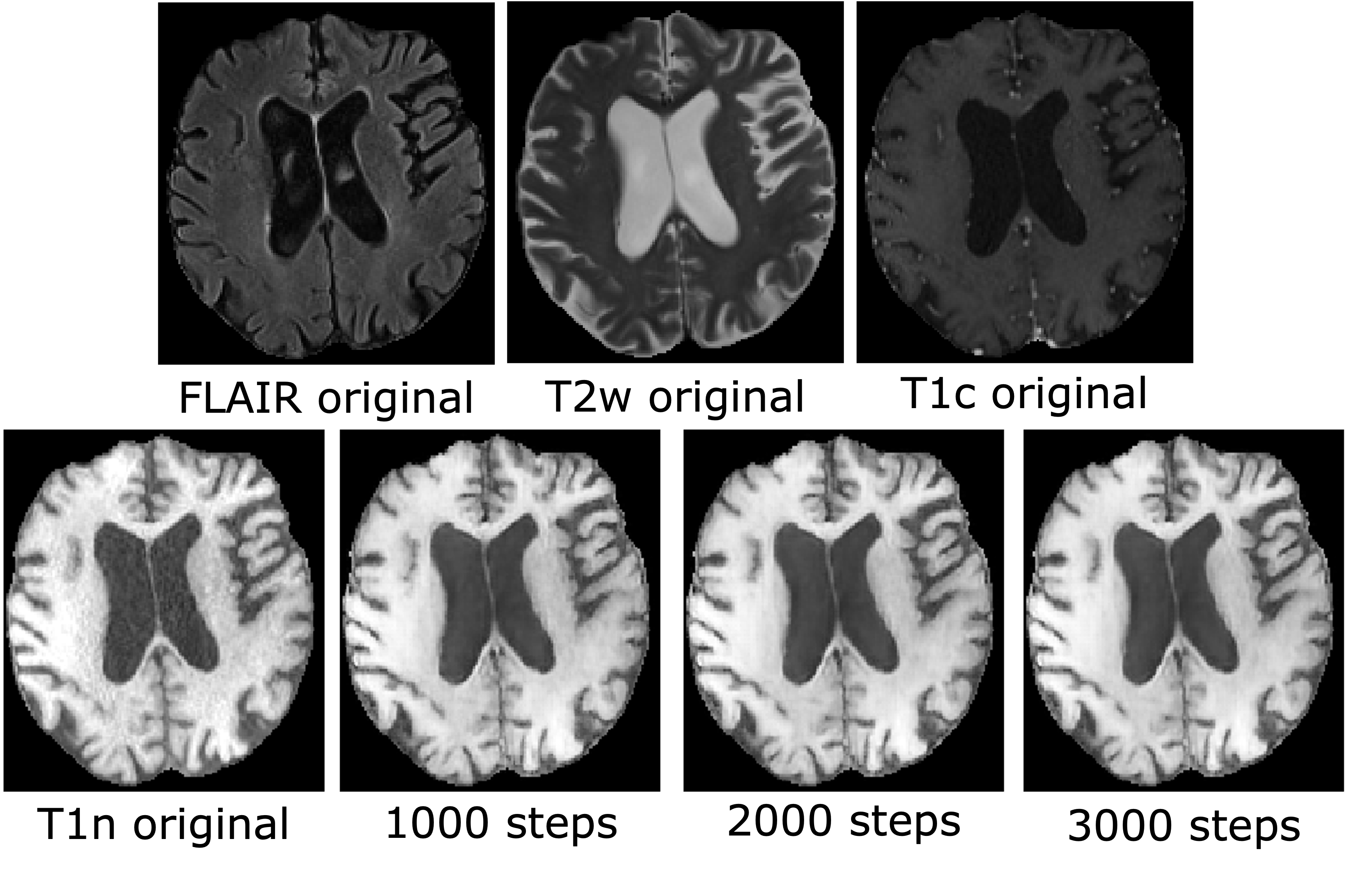}
\caption{Effect of the inference steps on the generated T1n modality of case 01301-000, knowing the FLAIR, T2w and T1c modalities.} \label{fig:Task7_visual}
\end{figure}

\begin{figure}
\includegraphics[width=\textwidth]{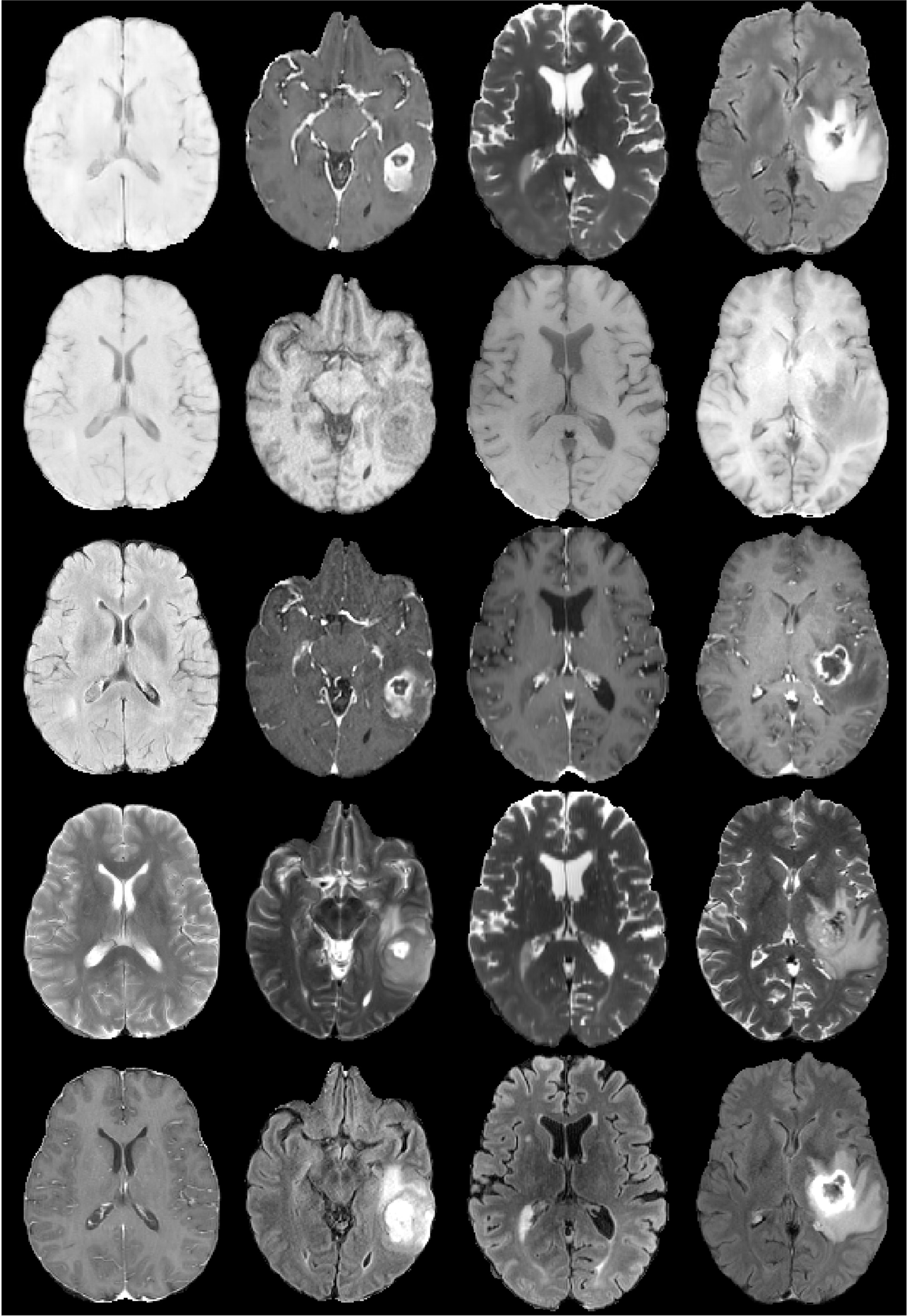}
\caption{Task 7 results. The first line shows the synthetically generated scan and the next 4 rows show the original modalities T1n, T1c, T2w and FLAIR. The generated scan in the first column is T1n (id 00008-001), the second T1c (id 00237-000), the third T2w (id 00078-000) and the last FLAIR (id 00469-001).} \label{fig:Visual_results_task7}
\end{figure}

For the missing modality task, only visual inspection was performed. Figure \ref{fig:Task7_visual} shows the case 01301-000 sampled using distinct number of sampling steps, and the \ref{fig:Visual_results_task7} show the results of several training cases using the model $K3T1_g$. The remaining models trained for task 7 did not produce good results, therefore we do not show any results here.

\section{Discussion}

Table \ref{tab:results_task8} shows that the model with the best performance is the $AKH_l$. As mentioned in \cite{lugmayr2022repaint}, increasing the number of steps, reducing the distance between each step, helps to improve the results of the generation, as can be seen in this table. Our best solution therefore uses 5000 steps to generate each case. The main disadvantage of using more sampling steps is the time required for each case. In our best solution, each case takes about 30 minutes. We also performed tests using different sampling schedulers, including repainting \cite{lugmayr2022repaint}. However, it produced very bad results.

When visually analysing the results of the $D_l$ and  $DC_l$ model, it was found that the model also learned to generate the tumours and not just the healthy tissue due to the amount of tumours present in the dataset. For this reason, the conditional models, especially the model without unhealthy tissue, performed better. In Figure \ref{fig:task_8_results} it can be seen that the presented solution shows good performance in replacing tumours with healthy tissue, however it is slightly blurred and, in the images that contain some artefacts, the overpainted region does not contain the same pattern of visual artefacts, which may have reduced the performance of the quantitative metrics. 

When comparing the reconstructions with real cases it is clear that the model has generated some hallucinations, which may be the reason why our solution ranked second place in the testing phase. Even though it looks realistic for someone with less expertise, it is not morphologically correct, and these high discrepancies of intensity increased the MSE. It is interesting to note that the PSNR and SSIM in the testing phase (Table \ref{tab:results_task8_test}) improved compared to the validation phase. This suggests that our solution is capable of generalizing well to unseen cases.

It is also important to note that our best model did not train long enough (did not converge), as we performed the evaluation in several checkpoints of the training and the results improved, but less and less improvements each time. Therefore, we believe that a longer training would improve our best solution. $D_l$ was the only model that converged.  

In Figures \ref{fig:Task7_visual} and \ref{fig:Visual_results_task7} it can be seen that the model can realistically generate the missing modality. Similar to the inpainting task, we tested the use of more sampling steps to improve the quality of the generated scans. All sampling strategies lead to very similar results, but it can be seen that more steps lead to less noisy results, i.e. smoother scans. We used 3000 steps in our submission. Each 1000 iterations takes roughly 7 minutes. From Table \ref{tab:results_task7_test}, it can be observed that the ET and TC regions were not well reconstructed, as it underperformed compared to the results achieved using the vanilla nnUNet \cite{isensee2021nnu}. The vanilla nnUNet achieved DSC scores of 82.03, 85.06, and 88.95, and HD95 values of 17.805, 17.337, and 8.498 for the ET, TC, and WT regions, respectively. The WT region demonstrated the best performance, although the results were still lower compared to those obtained using the nnUNet with real cases.

Future work should explore strategies such as the use of adversarial loss or consistency models to reduce the number of sampling steps required to achieve this performance. The reduction of sampling steps can also be achieved by using more efficient schedulers e.g., \textit{DPM++ 2M Karras}.

\begin{credits}
\subsubsection{\ackname} 
André Ferreira thanks the Fundação para a Ciência e Tecnologia (FCT) Portugal for the grant 2022.11928.BD. This work was supported by FCT within the R\&D Units Project Scope: UIDB/00319/2020. It was also supported by the Advanced Research Opportunities Program (AROP) of RWTH Aachen University and Clinician Scientist Program of the Faculty of Medicine RWTH Aachen University, also some of the computations were performed with computing resources granted by RWTH Aachen University under project 
rwth1484. This work received funding from enFaced (FWF KLI 678), enFaced 2.0 (FWF KLI 1044) and KITE (Plattform für KI-Translation Essen) from the REACT-EU initiative (EFRE-0801977, \url{https://kite.ikim.nrw/}).

\subsubsection{\discintname}
The authors have no competing interests to declare that are
relevant to the content of this article.
\end{credits}

\bibliographystyle{splncs04}
\bibliography{bib}

\begin{thebibliography}{10}
\providecommand{\url}[1]{\texttt{#1}}
\providecommand{\urlprefix}{URL }
\providecommand{\doi}[1]{https://doi.org/#1}

\bibitem{alexis2024diffusion}
Alexis, K., Christodoulidis, S., Gunopulos, D., Vakalopoulou, M.: Diffusion models for nuclei segmentation in low data regimes. In: IEEE International Symposium on Biomedical Imaging (2024)

\bibitem{baid2021rsna}
Baid, U., Ghodasara, S., Mohan, S., Bilello, M., Calabrese, E., Colak, E., Farahani, K., Kalpathy-Cramer, J., Kitamura, F.C., Pati, S., et~al.: The rsna-asnr-miccai brats 2021 benchmark on brain tumor segmentation and radiogenomic classification. arXiv preprint arXiv:2107.02314  (2021)

\bibitem{bakas2017advancing}
Bakas, S., Akbari, H., Sotiras, A., Bilello, M., Rozycki, M., Kirby, J.S., Freymann, J.B., Farahani, K., Davatzikos, C.: Advancing the cancer genome atlas glioma mri collections with expert segmentation labels and radiomic features. Scientific data  \textbf{4}(1),  1--13 (2017)

\bibitem{bakas2018identifying}
Bakas, S., Reyes, M., Jakab, A., Bauer, S., Rempfler, M., Crimi, A., Shinohara, R., Berger, C., Ha, S., Rozycki, M., et~al.: Identifying the best machine learning algorithms for brain tumor segmentation. progression assessment, and overall survival prediction in the BRATS challenge  \textbf{10} (2018)

\bibitem{baltruschat2024brasyn}
Baltruschat, I.M., Janbakhshi, P., Lenga, M.: Brasyn 2023 challenge: Missing mri synthesis and the effect of different learning objectives. arXiv preprint arXiv:2403.07800  (2024)

\bibitem{bieder2023memory}
Bieder, F., Wolleb, J., Durrer, A., Sandkuehler, R., Cattin, P.C.: Memory-efficient 3d denoising diffusion models for medical image processing. In: Medical Imaging with Deep Learning (2023)

\bibitem{dalmaz2022resvit}
Dalmaz, O., Yurt, M., {\c{C}}ukur, T.: Resvit: residual vision transformers for multimodal medical image synthesis. IEEE Transactions on Medical Imaging  \textbf{41}(10),  2598--2614 (2022)

\bibitem{dar2019image}
Dar, S.U., Yurt, M., Karacan, L., Erdem, A., Erdem, E., Cukur, T.: Image synthesis in multi-contrast mri with conditional generative adversarial networks. IEEE transactions on medical imaging  \textbf{38}(10),  2375--2388 (2019)

\bibitem{dhariwal2021diffusion}
Dhariwal, P., Nichol, A.: Diffusion models beat gans on image synthesis. Advances in neural information processing systems  \textbf{34},  8780--8794 (2021)

\bibitem{durrer2024denoising_inpaint}
Durrer, A., Cattin, P.C., Wolleb, J.: Denoising diffusion models for inpainting of healthy brain tissue. arXiv preprint arXiv:2402.17307  (2024)

\bibitem{durrer2024denoising}
Durrer, A., Wolleb, J., Bieder, F., Friedrich, P., Melie-Garcia, L., Ocampo-Pineda, M., Bercea, C.I., Hamamci, I.E., Wiestler, B., Piraud, M., et~al.: Denoising diffusion models for 3d healthy brain tissue inpainting. arXiv preprint arXiv:2403.14499  (2024)

\bibitem{ezhov2021geometry}
Ezhov, I., Mot, T., Shit, S., Lipkova, J., Paetzold, J.C., Kofler, F., Pellegrini, C., Kollovieh, M., Navarro, F., Li, H., et~al.: Geometry-aware neural solver for fast bayesian calibration of brain tumor models. IEEE Transactions on Medical Imaging  \textbf{41}(5),  1269--1278 (2021)

\bibitem{ezhov2023learn}
Ezhov, I., Scibilia, K., Franitza, K., Steinbauer, F., Shit, S., Zimmer, L., Lipkova, J., Kofler, F., Paetzold, J.C., Canalini, L., et~al.: Learn-morph-infer: a new way of solving the inverse problem for brain tumor modeling. Medical Image Analysis  \textbf{83},  102672 (2023)

\bibitem{ferreira2022generation}
Ferreira, A., Magalh{\~a}es, R., M{\'e}riaux, S., Alves, V.: Generation of synthetic rat brain mri scans with a 3d enhanced alpha generative adversarial network. Applied Sciences  \textbf{12}(10), ~4844 (2022)

\bibitem{ferreira2024we}
Ferreira, A., Solak, N., Li, J., Dammann, P., Kleesiek, J., Alves, V., Egger, J.: How we won brats 2023 adult glioma challenge? just faking it! enhanced synthetic data augmentation and model ensemble for brain tumour segmentation. arXiv preprint arXiv:2402.17317  (2024)

\bibitem{friedrich2024wdm}
Friedrich, P., Wolleb, J., Bieder, F., Durrer, A., Cattin, P.C.: Wdm: 3d wavelet diffusion models for high-resolution medical image synthesis. arXiv preprint arXiv:2402.19043  (2024)

\bibitem{gsaxner2024deepdr}
Gsaxner, C., Mori, S., Schmalstieg, D., Egger, J., Paar, G., Bailer, W., Kalkofen, D.: Deepdr: Deep structure-aware rgb-d inpainting for diminished reality. In: 2024 International Conference on 3D Vision (3DV). pp. 750--760. IEEE (2024)

\bibitem{isensee2021nnu}
Isensee, F., J{\"a}ger, P.F., Full, P.M., Vollmuth, P., Maier-Hein, K.H.: nnu-net for brain tumor segmentation. In: Brainlesion: Glioma, Multiple Sclerosis, Stroke and Traumatic Brain Injuries: 6th International Workshop, BrainLes 2020, Held in Conjunction with MICCAI 2020, Lima, Peru, October 4, 2020, Revised Selected Papers, Part II 6. pp. 118--132. Springer (2021)

\bibitem{isola2017image}
Isola, P., Zhu, J.Y., Zhou, T., Efros, A.A.: Image-to-image translation with conditional adversarial networks. In: Proceedings of the IEEE conference on computer vision and pattern recognition. pp. 1125--1134 (2017)

\bibitem{khader2022medical}
Khader, F., Mueller-Franzes, G., Arasteh, S.T., Han, T., Haarburger, C., Schulze-Hagen, M., Schad, P., Engelhardt, S., Baessler, B., Foersch, S., et~al.: Medical diffusion: denoising diffusion probabilistic models for 3d medical image generation. arXiv preprint arXiv:2211.03364  (2022)

\bibitem{kofler2023brain}
Kofler, F., Meissen, F., Steinbauer, F., Graf, R., Oswald, E., de~da Rosa, E., Li, H.B., Baid, U., Hoelzl, F., Turgut, O., et~al.: The brain tumor segmentation (brats) challenge 2023: Local synthesis of healthy brain tissue via inpainting. arXiv preprint arXiv:2305.08992  (2023)

\bibitem{li2023brain}
Li, H.B., Conte, G.M., Anwar, S.M., Kofler, F., Ezhov, I., van Leemput, K., Piraud, M., Diaz, M., Cole, B., Calabrese, E., et~al.: The brain tumor segmentation (brats) challenge 2023: Brain mr image synthesis for tumor segmentation (brasyn). ArXiv  (2023)

\bibitem{lugmayr2022repaint}
Lugmayr, A., Danelljan, M., Romero, A., Yu, F., Timofte, R., Van~Gool, L.: Repaint: Inpainting using denoising diffusion probabilistic models. In: Proceedings of the IEEE/CVF conference on computer vision and pattern recognition. pp. 11461--11471 (2022)

\bibitem{menze2014multimodal}
Menze, B.H., Jakab, A., Bauer, S., Kalpathy-Cramer, J., Farahani, K., Kirby, J., Burren, Y., Porz, N., Slotboom, J., Wiest, R., et~al.: The multimodal brain tumor image segmentation benchmark (brats). IEEE transactions on medical imaging  \textbf{34}(10),  1993--2024 (2014)

\bibitem{nichol2021improved}
Nichol, A.Q., Dhariwal, P.: Improved denoising diffusion probabilistic models. In: International conference on machine learning. pp. 8162--8171. PMLR (2021)

\bibitem{saharia2022palette}
Saharia, C., Chan, W., Chang, H., Lee, C., Ho, J., Salimans, T., Fleet, D., Norouzi, M.: Palette: Image-to-image diffusion models. In: ACM SIGGRAPH 2022 conference proceedings. pp. 1--10 (2022)

\bibitem{xiao2021tackling}
Xiao, Z., Kreis, K., Vahdat, A.: Tackling the generative learning trilemma with denoising diffusion gans. arXiv preprint arXiv:2112.07804  (2021)

\bibitem{yu2019ea}
Yu, B., Zhou, L., Wang, L., Shi, Y., Fripp, J., Bourgeat, P.: Ea-gans: edge-aware generative adversarial networks for cross-modality mr image synthesis. IEEE transactions on medical imaging  \textbf{38}(7),  1750--1762 (2019)

\bibitem{zhu2023make}
Zhu, L., Xue, Z., Jin, Z., Liu, X., He, J., Liu, Z., Yu, L.: Make-a-volume: Leveraging latent diffusion models for cross-modality 3d brain mri synthesis. In: International Conference on Medical Image Computing and Computer-Assisted Intervention. pp. 592--601. Springer (2023)

\bibitem{zhu2024advancing}
Zhu, R., Zhang, X., Pang, H., Xu, C., Ye, C.: Advancing brain tumor inpainting with generative models. arXiv preprint arXiv:2402.01509  (2024)

\end{thebibliography}
\end{document}